\documentclass[letterpaper]{article} 
\usepackage{aaai2026}  
\usepackage{times}  
\usepackage{helvet}  
\usepackage{courier}  
\usepackage[hyphens]{url}  
\usepackage{graphicx} 
\usepackage{natbib}  
\usepackage{caption} 
\usepackage{algorithm}
\usepackage{algorithmic}

\usepackage{booktabs}
\usepackage[utf8]{inputenc}  
\usepackage[T1]{fontenc}
\usepackage[english]{babel}
\usepackage{float} 
\usepackage{multirow}
\usepackage{colortbl}
\usepackage{hhline}
\usepackage{enumitem}
\usepackage{booktabs}
\usepackage{makecell}
\usepackage{tabularx}        
\usepackage{graphicx}        
\usepackage{xcolor}

\usepackage{rotating}   
\usepackage{amssymb}    
\usepackage{pifont}     
\usepackage{graphicx}

\newcommand{\cmark}{\checkmark}
\newcommand{\xmark}{\ding{55}}

\usepackage{newfloat}
\usepackage{listings}
\DeclareCaptionStyle{ruled}{labelfont=normalfont,labelsep=colon,strut=off} 
\floatstyle{ruled}
\newfloat{listing}{tb}{lst}{}
\floatname{listing}{Listing}
\title{
    CookBench: A Long-Horizon Embodied Planning Benchmark for Complex Cooking Scenarios}
\author{
    Muzhen Cai\textsuperscript{\rm 1},
    Xiubo Chen\textsuperscript{\rm 1},
    Yining An\textsuperscript{\rm 1},
    Jiaxin Zhang\textsuperscript{\rm 1},
    Xuesong Wang\textsuperscript{\rm 1},
    Wang Xu\textsuperscript{\rm 2},
    Weinan Zhang\textsuperscript{\rm 1},
    Ting Liu\textsuperscript{\rm 1}
}
\affiliations{
    \textsuperscript{\rm 1}Harbin Institute of Technology\\
    \textsuperscript{\rm 2}Tsinghua University\\
    mzcai@ir.hit.edu.cn, 2022111792@stu.hit.edu.cn, 2022113080@stu.hit.edu.cn, 120L020416@stu.hit.edu.cn, xswang@ir.hit.edu.cn, xwjim812@gmail.com, wnzhang@ir.hit.edu.cn, tliu@ir.hit.edu.cn
}

\usepackage{bibentry}

\begin{document}

\maketitle


\begin{abstract}

Embodied Planning is dedicated to the goal of creating agents capable of executing long-horizon tasks in complex physical worlds. However, existing embodied planning benchmarks frequently feature short-horizon tasks and coarse-grained action primitives. To address this challenge, we introduce \textbf{CookBench}, a benchmark for long-horizon planning in complex cooking scenarios. By leveraging a high-fidelity simulation environment built upon the powerful Unity game engine, we define frontier AI challenges in a complex, realistic environment. The core task in CookBench is designed as a two-stage process. First, in \textbf{Intention Recognition}, an agent needs to accurately parse a user's complex intent. Second, in \textbf{Embodied Interaction}, the agent should execute the identified cooking goal through a long-horizon, fine-grained sequence of physical actions. Unlike existing embodied planning benchmarks, we refine the action granularity to a spatial level that considers crucial operational information while abstracting away low-level robotic control. Besides, We provide a comprehensive toolset that encapsulates the simulator. Its unified API supports both macro-level operations, such as placing orders and purchasing ingredients, and a rich set of fine-grained embodied actions for physical interaction, enabling researchers to focus on high-level planning and decision-making. Furthermore, we present an in-depth analysis of state-of-the-art, closed-source Large Language Model and Vision-Language Model, revealing their major shortcomings and challenges posed by complex, long-horizon tasks. The full benchmark will be open-sourced to facilitate future research.

\end{abstract}

\section{Introduction}
\label{intro}

\begin{figure}[t]
  \centering
  \includegraphics[width=\columnwidth]{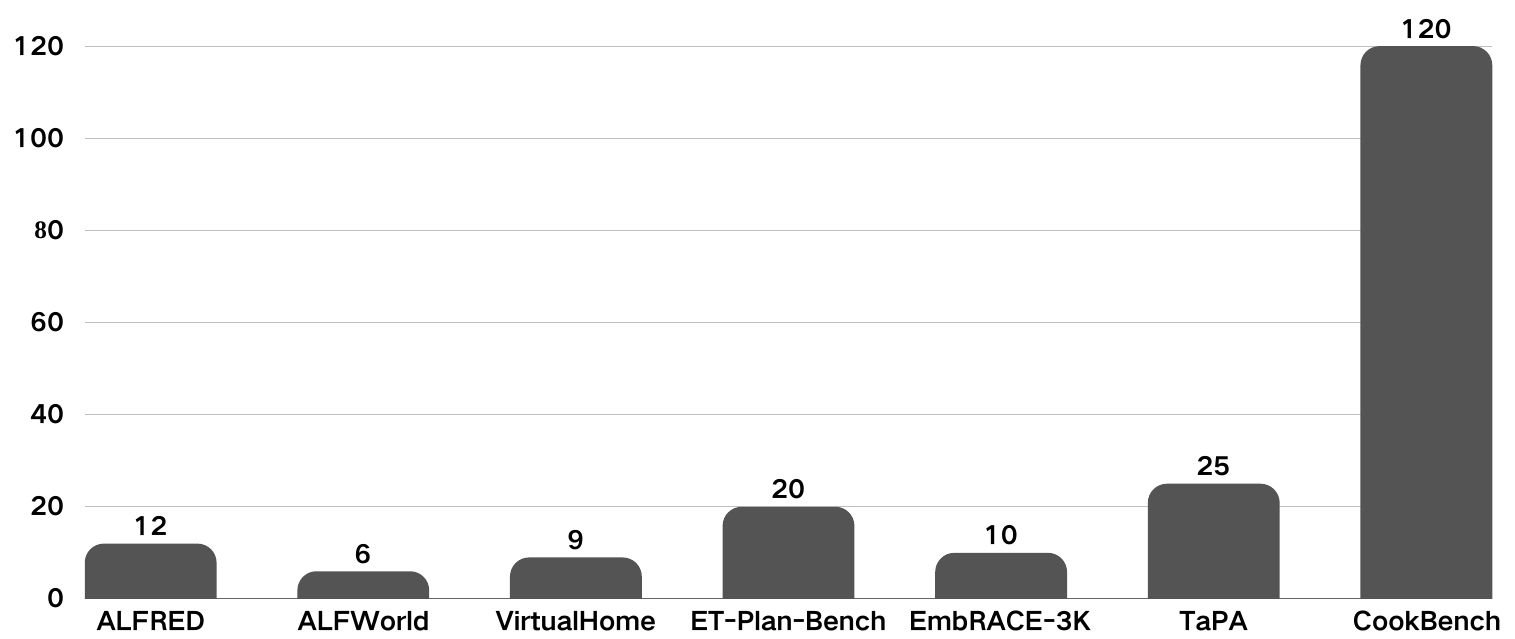}
  \caption{A Long-Horizon Comparison of Embodied Planning Benchmarks.}
  \label{fig:tab-bench}
\end{figure}

The core vision of Embodied AI~\cite{liu2024aligning, duan2022survey} is to create agents capable of perceiving, reasoning, and acting autonomously within complex, dynamic physical worlds. Under this vision, Embodied Planning~\cite{song2023llm}, as a crucial subfield, is dedicated to investigating how agents can autonomously decompose complex goals into a sequence of actions to interact with the physical world, thereby completing long-horizon, multi-step tasks.

However, progress in the field of embodied planning~\cite{kim2024realfred} is largely guided and constrained by existing benchmarks. Current mainstream benchmarks, such as Embodied-Bench~\cite{yang2025embodiedbench}, VirtualHome~\cite{puig2018virtualhome}, TEACh~\cite{padmakumar2022teach}, ALFRED~\cite{shridhar2020alfred}, and ET-Plan-bench~\cite{zhang2024plan}, while having made significant contributions, share a core limitation: focusing on short-to-medium-horizon tasks~\cite{wu2023embodied}, as shown in Figure~\ref{fig:tab-bench}. For example, the average task length in ALFRED is only around 12 steps, which is insufficient to rigorously test an agent's long-term memory, complex reasoning, and multi-task scheduling capabilities. This focus on shorter tasks often means they do not require deep domain knowledge or complex interaction logic~\cite{shridhar2020alfworld}. Consequently, there is a pressing need for a benchmark that can effectively evaluate an agent's capabilities in \textbf{long-horizon planning}~\cite{duan2022survey, feldotto2022deploying, kaur2023review}.

To bridge this gap in long-horizon planning research, we introduce \textbf{CookBench}, an embodied benchmark specifically designed for complex cooking scenarios. We choose cooking as the task domain because it naturally encompasses the key challenges of \textbf{long-horizon planning}~\cite{mavrogiannis2024cook2ltl, huang2022inner}, \textbf{multitasking}, \textbf{irreversible state transformations}~\cite{bollini2013interpreting}, and a deep reliance on \textbf{domain knowledgee}~\cite{kanazawa2024real, ivanova2025ambik}. The core task in CookBench is designed as a two-stage process. First, in Intention Recognition, an agent must parse complex user intent from a dataset we constructed of approximately 14,394 bilingual (Chinese and English) natural language instructions. Second, in Embodied Interaction, the agent should execute the identified cooking goal within a challenging suite of 131 single-dish and 4,446 multi-dish tasks by performing a long-horizon, fine-grained sequence of physical actions to complete the dish preparation. The CookBench simulation environment is built upon the high-fidelity engine Unity~\cite{haas2014history}.


On this platform, we conduct experiments with a representative set of powerful language and vision-language model. 
Our embodied interaction stage requires agents to plan from purely visual input, tackling an integrated challenge of navigation, manipulation, and other complex behaviors.
As no existing work can currently solve such a comprehensive task autonomously, we introduce a human-in-the-loop (HITL) approach for this initial exploratory evaluation.
Our analysis shows that while these leading models exhibit certain planning capabilities, the long-horizon challenges of CookBench still pose a severe test. Through a systematic analysis of their performance, we identify critical shortcomings in areas such as long-horizon state tracking and physical commonsense reasoning.
These findings are not a declaration of failure, but rather illuminate the core technical directions that must be addressed to build more capable embodied agents for the a future.

The main contributions of this paper are as follows:
\begin{itemize}
    \item \textbf{A Novel Benchmark for Long-Horizon Planning.} We introduce CookBench, a complex, knowledge-driven benchmark with an average task length of ~120 steps, designed to rigorously evaluate long-horizon planning capabilities.

    \item \textbf{A Large-Scale Dataset and Task Suite.} We provide a comprehensive, bilingual dataset with 14,394 instructions for complex intent recognition, alongside a challenging suite of embodied interaction tasks, including 131 single-dish and approximately 4,446 multi-dish scenarios.

    \item \textbf{An Empirical Evaluation and Bottleneck Analysis of Leading Models.} We present a human-in-the-loop feasibility study that not only validates CookBench as an operational and evaluable testbed but also provides a systematic analysis of the critical bottlenecks of leading models in areas like long-term memory and scheduling.
\end{itemize}


\begin{table*}[t]
    \centering
    \small
    \begin{tabular}{@{}l l cc cccc ccccc@{}}
        \toprule
        \multicolumn{2}{c}{\multirow{2}{*}{\textbf{Benchmark}}} & \multicolumn{2}{c}{\textbf{- Task \& Eval. -}} & \multicolumn{4}{c}{\textbf{--- Environment \& Interaction ---}} & \multicolumn{5}{c}{\textbf{--- Core Capabilities ---}} \\
        \cmidrule(lr){3-4} \cmidrule(lr){5-8} \cmidrule(lr){9-13}
        & & \makecell{\textbf{Seq.}\\\textbf{Len.}} & \makecell{\textbf{Auto.}\\\textbf{Eval.}} & \textbf{Simulator} & \textbf{\# Objs} & \textbf{\# Tools} & \makecell{\textbf{State}\\\textbf{Chg.}} & \textbf{Intent} & \textbf{Plan} & \textbf{Nav.} & \makecell{\textbf{Manip.}\\\textbf{Gran.}} & \textbf{Knowledge} \\
        \midrule
        
        \multirow{11}{*}{\rotatebox[origin=c]{90}{\makecell{Planning}}} 
        
        & ALFRED & \textasciitilde12 & \cmark & AI2-THOR & 112 & \xmark & Discrete & \xmark & \cmark & \xmark & Coarse & Common \\
        
        & ReALFRED & \textasciitilde12 & \cmark & AI2-THOR & 112 & \xmark & Discrete & \xmark & \cmark & \cmark & Coarse & Common \\
        
        & EB-ALFRED & \textasciitilde12 & \cmark & AI2-THOR & 112 & \xmark & Discrete & \xmark & \cmark & \xmark & Coarse & Common \\
        
        
        & ALFWorld & \textasciitilde6 & \cmark & Text-only & 66 & \xmark & Discrete & \xmark & \cmark & \xmark & Coarse & Common \\
        
        & VirtualHome & \textasciitilde9 & \cmark & Unity3D & 349 & \xmark & Discrete & \xmark & \cmark & \xmark & Coarse & Common \\

        & ET-Plan-bench & <20 & \cmark & \makecell{Unity3D \\\& Habitat} & >300 & \xmark & Discrete & \cmark & \cmark & \xmark & Coarse & Common \\
        
        & EmbRACE-3K & \textasciitilde10 & \cmark & UnrealCV & --- & \xmark & Discrete & \xmark & \cmark & \cmark & Coarse & Common \\

        & TEACh & --- & \cmark & AI2-THOR & 104 & \xmark & Discrete & \xmark & \cmark & \xmark & Coarse & Common \\

        & ProcTHOR & --- & \cmark & ProcTHOR & --- & \xmark & Discrete & \xmark & \cmark & \cmark & --- & Common \\
        
        & TaPA & \textasciitilde25 & \xmark & --- & --- & \xmark & Discrete & \xmark & \cmark & \xmark & --- & Common \\
        \cmidrule(lr){2-13}

        \multirow{2}{*}{\rotatebox[origin=c]{90}{\makecell{Manip.}}}

        & BEHAVIOR & --- & \cmark & iGibson & 391 & \xmark & Discrete & \xmark & \cmark & \cmark & Fine  & Common \\

        & ManiSkill2 & --- & \cmark & SAPIEN & --- & \xmark & Discrete & \xmark & \xmark & \xmark & Fine  & \makecell{Common} \\

        \cmidrule(lr){2-13}
        & \textbf{CookBench} & \textasciitilde120 & \cmark &Unity & 430 & 40 & \makecell{Conti-\\nuous} & \cmark & \cmark & \cmark & Fine & \makecell{Common\\+Domain\\+Tool} \\

        \bottomrule
    \end{tabular}
    \caption{A comprehensive comparison of major embodied AI benchmarks, including Embodied Planning and Embodied Manipulation. Key column definitions are as follows: 
    \textbf{Seq. Len.}: Average number of high-level action steps per task. 
    \textbf{\# Objs / \# Tools}: The count of interactive objects and tools. We define a ``tool'' by its function: an item is only a tool if it can execute a specific action on another object. For example, a knife that can cut is a tool; one that can only be picked up and placed is not.
    \textbf{State Chg.}: Nature of object state changes. 
    \textbf{Intent}: The need to infer latent goals from ambiguous or interactive instructions.
    \textbf{Plan}: The requirement for multi-step, task-level planning. 
    \textbf{Nav.}: The necessity for the agent to perform low-level navigation actions.
    \textbf{Manip. Gran.} refers to manipulation granularity.
    \textbf{Knowledge}: The type of knowledge required.
    }
    \label{tab:compare_bench} 
\end{table*}

\section{Related Works}
\label{sec:related_work}

Long-horizon planning is a cornerstone of autonomous behavior in Embodied AI~\cite{liu2025embodied}, requiring agents to execute extended sequences of actions to achieve complex goals~\cite{yang2025magma}. This capability poses several fundamental challenges not present in short-horizon tasks, including the need for robust \textbf{long-term memory} to track object states and task progress, \textbf{hierarchical reasoning} to decompose high-level goals~\cite{liu2024aligning}, and effective \textbf{multi-task scheduling} to manage parallel processes and shared resources. To systematically analyze how existing benchmarks address these long-horizon challenges, we categorize them into three types based on the nature and length of their tasks, which is shown in Table~\ref{tab:compare_bench}.

\paragraph{Short-Horizon Manipulation Benchmarks.} 
The core challenge of short-horizon benchmarks lies in low-level, fine-grained physical control\cite{zhang2025generative}. This domain is a primary focus of current research in embodied manipulation, where the availability of high-fidelity simulation environments greatly facilitate extensive exploration. Benchmarks such as BEHAVIOR~\cite{srivastava2022behavior} and ManiSkill2~\cite{gu2023maniskill2}, built upon physics engines like MuJoCo~\cite{todorov2012mujoco}, PyBullet\cite{yang2021open}, iGibson~\cite{li2021igibson}, and SAPIEN~\cite{xiang2020sapien}, achieve remarkable realism in simulating dexterous manipulation. This includes modeling complex forward and inverse kinematics, contact-rich dynamics, and friction. While these benchmarks are essential for robotics research, their focus is on the low-level implementation details of atomic actions—such as how to adjust a robot arm’s pose to grasp an object—but they lack the high-level planning of the steps required to complete a task\cite{li2025developments}. As a comprehensive review of these benchmarks is beyond the scope of this paper, we list a representative selection in Table~\ref{tab:compare_bench} for illustrative comparison.

\paragraph{Short-Horizon Sequential Task Benchmarks.}
This category represents the mainstream of current embodied planning research, with task lengths typically ranging from 10-30 steps. In the vision-language domain, the ALFRED series (ALFRED~\cite{shridhar2020alfred}, ReALFRED~\cite{kim2024realfred}, EB-ALFRED~\cite{yang2025embodiedbench}) is the pioneering work, requiring an agent to follow instructions in 3D visual environments and execute a standardized, multi-step plan by identifying the location of target objects (e.g., via segmentation masks)\cite{chen2024spatialvlm}. Building on this paradigm, ProcTHOR~\cite{deitke2022procthor} focuses on generalization in procedurally generated scenes, while TEACh~\cite{padmakumar2022teach} and EmbRACE-3K~\cite{lin2025embrace} explore interactive planning through human-agent dialogue.

In the symbolic-abstract domain, researchers focuses the challenge entirely on logical task decomposition and symbolic reasoning by abstracting the world into symbols or text~\cite{jin2023mini}. VirtualHome~\cite{puig2018virtualhome} generates activities via programmatic scripts, while ALFWorld~\cite{shridhar2020alfworld} translates the ALFRED tasks into a text-based game. ET-Plan-bench~\cite{zhang2024plan} and TaPA~\cite{wu2023embodied} push this direction further, presenting more complex and longer-horizon planning puzzles at a logical level by moving away from or simplifying the constraints of a full simulation environment. While this approach increases logical complexity and length, it comes at the cost of sacrificing the core challenges of embodied planning. The agent is no longer required to ground its plan in a limited set of atomic actions supported by the environment~\cite{liang2022code, zhu2025move}, nor is it tested on its ability to reason with complex scene information and physical commonsense~\cite{zhou2025roborefer, xu20253d}.

Despite their significant contributions, these benchmarks share common limitations from a long-horizon planning perspective. First, their task lengths are insufficient to stress-test an agent's long-term memory and strategic capabilities. Second, the task flows are often linear, primarily testing the ability to follow a sequence rather than requiring multi-task scheduling and optimization. Finally, their interaction action sets are relatively finite and abstract, simplifying the complex physics and tool-use inherent in the real world. This makes them ill-suited for more complex tasks, such as cooking, that demand long-horizon strategic planning.

\paragraph{The Gap: Long-Horizon Strategic Planning.}
Long-horizon strategic planning represents a critical research gap. Benchmarks in this category would need to feature ultra-long sequences (>100 steps) with complex, multi-threaded workflows, resource contention, and a deep reliance on domain knowledge. Currently, the field lacks a platform that systematically provides and evaluates these challenges. \textbf{CookBench} is designed to directly fill this gap.
\section{The CookBench Environment and Task Framework}

This section introduces the composition of CookBench in detail, which consists of a high-fidelity simulation platform, a decoupled task framework, a structured knowledge base of 131 dishes and products and tools, and corresponding intelligent agent interfaces and evaluation system.

\begin{figure*}[t]
    \centering 
    \includegraphics[width=\textwidth]{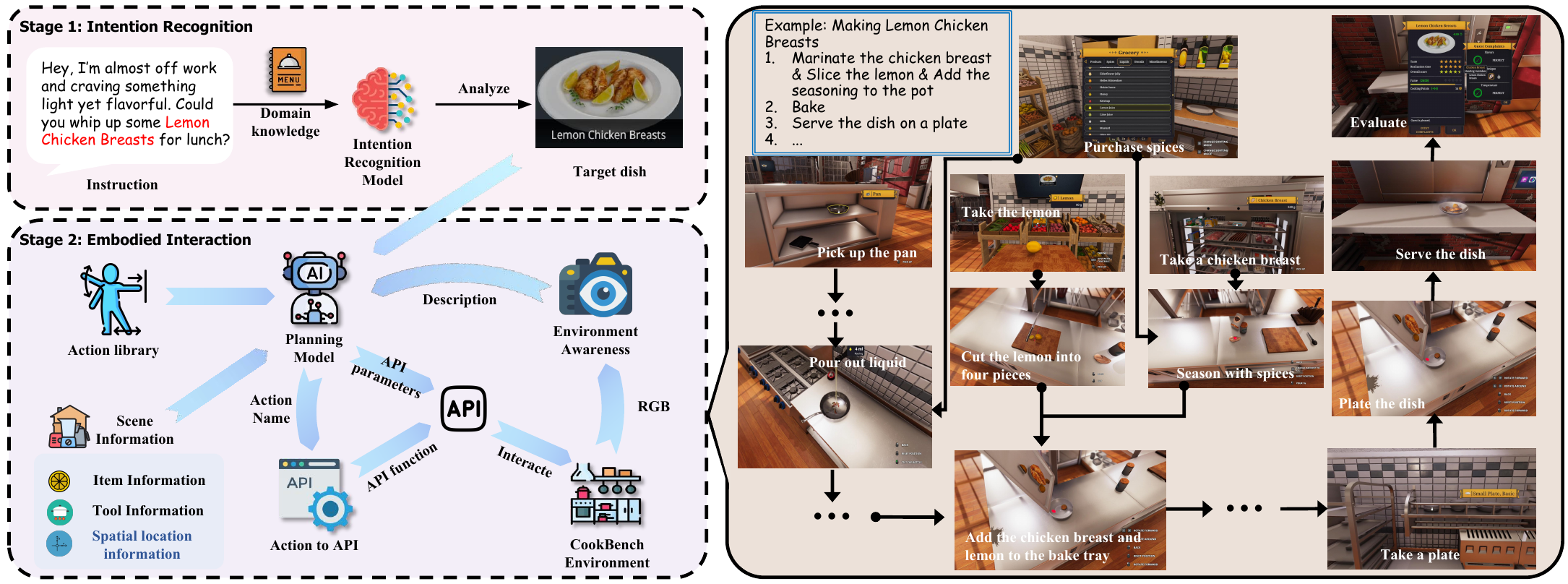}
    \caption{An overview of the CookBench framework and an example task dependency graph. The framework (left) uses a two-stage process: Intention Recognition to parse natural language into a concrete goal, and Embodied Interaction for execution. In the interaction stage, a planning model must predict both the action function to call from an API (e.g., move\_vertical) and its specific operational parameters (e.g., distance\_pixels = -300). The example workflow (right) is shown as a dependency graph whose parallel structure challenges the agent to plan and optimize for efficiency.}
    \label{fig:cookbench_pipeline} 
\end{figure*}

\subsection{A High-Fidelity Simulation Platform}
\begin{figure*}[t]
    \centering 
    \includegraphics[width=\textwidth]{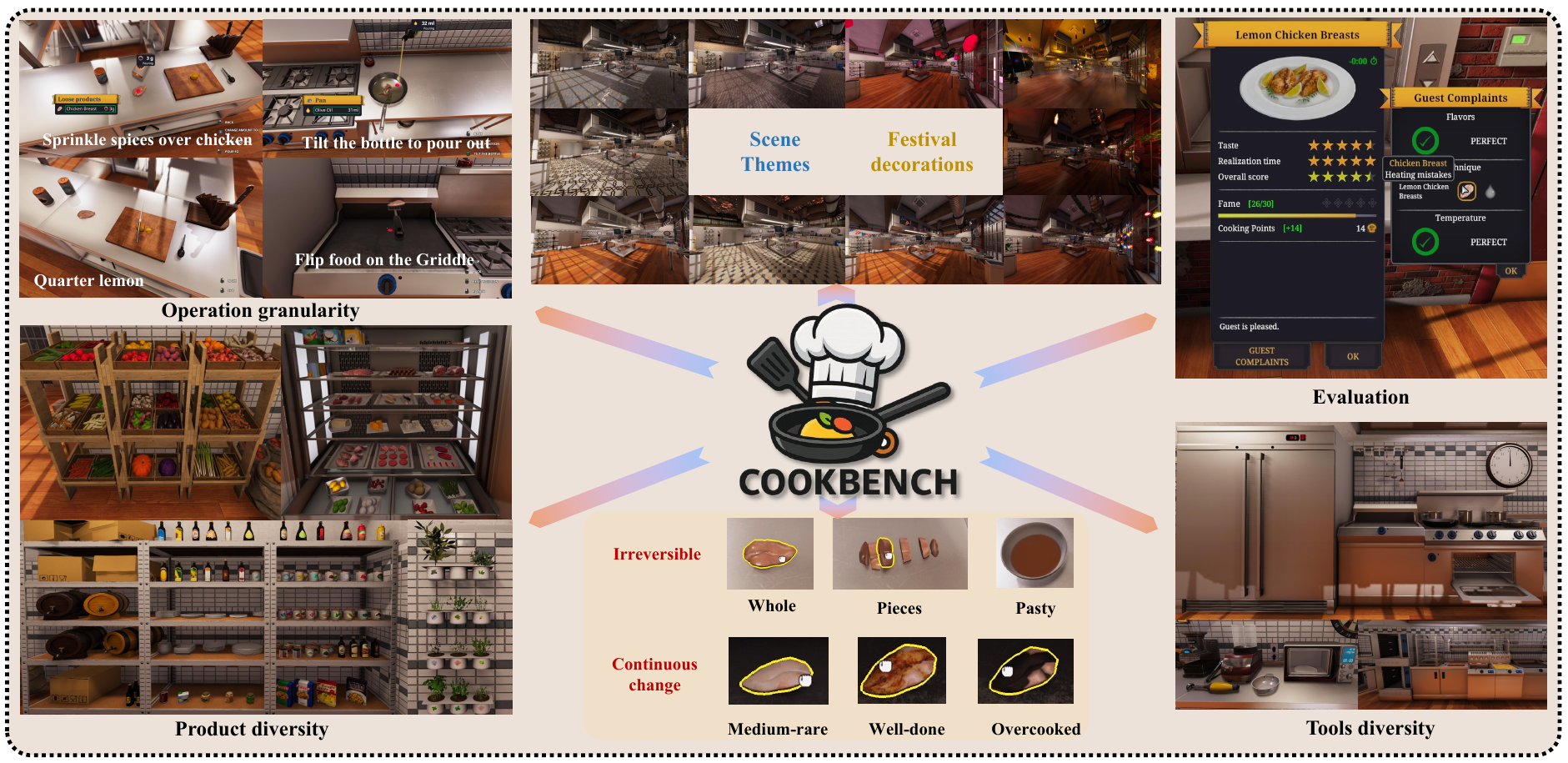}
    \caption{An overview of the CookBench simulation environment, highlighting its key features: extensive diversity in products and tools; fine-grained interactions that produce complex state changes (e.g., irreversible and continuous); high customizability with 10 scene themes and 10 festival decorations (five of each are shown for brevity); and a built-in, automated evaluation system.}
    \label{fig:cookbench_scene} 
\end{figure*}

CookBench is built upon \textit{Cooking Simulator}, a high-fidelity simulation environment powered by the Unity physics simulation engine~\cite{haas2014history}, ensuring high-fidelity visual effects and physics-based feedback. All tasks within CookBench are situated in the \textbf{Sandbox Mode} of the \textbf{Classic} kitchen scene. This simulator not only achieves photorealistic visuals but, more importantly, provides a dynamic world with consistent physical laws and rich interaction. The environment is extensively equipped, featuring over \textbf{228 food items}, \textbf{132 liquids}, \textbf{70 spices}, and a vast array of tools, as shown in Figure~\ref{fig:cookbench_scene}.

To support large-scale and diverse experiments, our benchmark is built around a knowledge base that specifies \textbf{131 recipes}. Each entry contains the dish's name, its required ingredients, and a procedural recipe written in a compact, semi-structured text format that requires intelligent parsing. For example, the recipe for ``Lemon Chicken Breasts'' is: \textit{Add to a Pot: Olive Oil [50ml], White Wine [10ml], Lemon Juice [10ml], Oregano, dried [5g], Thyme Twig [10g], Garlic [60g]. Fry for 60s. Take Chicken Breast [240g]. Season with: Salt [5g], Black Pepper [5g]. Cut Lemon [80g] into Pieces [16g]. Add to a Baking Tray: Mixture [70ml], Chicken Breast, Lemon. Fry for 90s. Transfer into a Deep Plate: Mixture [70ml], Chicken Breast, Lemon. Serve hot}.

To facilitate subsequent structured analysis and task generation, these dishes are systematically classified along two independent dimensions. The first dimension is 'difficulty of preparation', where dishes are classified into three levels based on criteria like procedural complexity and the number of sub-tasks: simple (31 dishes), medium (72 dishes), and hard (28 dishes). The second dimension is 'cuisine/regional style', where dishes are categorized into six major types by analyzing their names and signature ingredients. These include French \& Mediterranean cuisine (32 dishes) and American cuisine (26 dishes). More details are listed in the appendix. 

\subsection{A Decoupled Two-Layer Task Framework}
The core task framework of CookBench is designed as a clearly decoupled two-layer structure, separately addressing the challenges of intention recognization and physical execution.

\subsubsection{Task Stage 1: Natural Language Intent Recognition.}
The objective of this layer is to test an agent's ability to accurately infer a user's final cooking intent from complex natural language instructions. For this, we have systematically designed bilingual (Chinese/English) test sets covering both single-dish and multi-dish scenarios, with approximately 7,197 instructions per language, named \textbf{CookBench-IR}. The task requires the agent to predict the final list of target dishes from the instruction; for multi-dish instructions, the model should also parse the correct dish sequence if one is specified by the user. 



Single-dish instructions are designed to evaluate the model's robustness in fine-grained semantic understanding and are divided into three difficulty levels. \textbf{Direct Instructions} are simple commands with a clear dish name, testing basic named entity recognition. \textbf{Interference Instructions} contain multiple options or conflicting information (covering seven realistic scenarios like ``Referential Resolution'' and ``Logical Branching''), requiring the model to perform logical reasoning to identify the final choice. Lastly, \textbf{Ambiguous Instructions} lack a dish name, challenging the agent to infer the dish from vague descriptions of ingredients, taste, or even domain knowledge (e.g., dietary therapy, cuisine) and testing its ability to perform deep reasoning and ground fuzzy language to a structured knowledge base.

To comprehensively evaluate planning capabilities, our multi-dish instruction set is structurally balanced across several dimensions. Instructions are distributed across scenarios of 4, 5, and 6 dishes, with approximately 1,482 instances per scenario. Furthermore, we carefully control the ratio of simple to interference-based instructions within each command and ensure a uniform distribution of dish mentions to prevent model bias. The dataset also covers different ordering patterns, such as \textbf{batch orders} (where all dishes are specified at once) and \textbf{sequential orders} (where dishes are specified incrementally). This distinction is crucial as it has significant downstream implications for the embodied interaction in Stage 2: batch orders allow for global planning and resource optimization, while sequential orders pose a more severe challenge to an agent's state tracking and dynamic scheduling capabilities.

\subsubsection{Task Stage 2: Embodied Interaction.}
This stage evaluates the agent's capacity to translate the structured intent from Task 1 into a concrete, long-horizon sequence of physical actions. The goal is to follow a recipe and successfully prepare one or more dishes in a dynamic, physics-based kitchen environment. The tasks consist of 131 single-dish and 4,446 multi-dish scenarios, consistent with the dish combinations in Stage 1. As shown in Figure~\ref{fig:cookbench_pipeline}, the planning model needs to devise the action names and parameters to complete the task, based on the atomic action library, perceptual information, and scene information.

The agent's interaction with the world is exclusively conducted through a predefined \textbf{Atomic Action Library}. This skill manual contains approximately 35 fundamental physical actions that define its complete range of capabilities. These actions cover a wide spectrum, from basic movement and perspective (e.g., Move Forward, Look Around, Crouch), to core object interactions (e.g., Pick Up, Put Down, Open Door, Pick Up with Tongs), fine-grained food processing (e.g., Cut, Stir / Mix, Crack Open, Separate), operations with cookware and liquids (e.g., Pour, Flip Over), and control over environmental appliances (e.g., Turn On Appliance, Start Heating, Add Time). All of these atomic actions are implemented via keyboard and mouse APIs.

The agent is provided with static \textbf{Scene Information} to inform its planning. 
This ``world map'' includes detailed information about objects and tools (e.g., their weight, initial relative positions, APIs, and supported operations) as well as spatial relationships derived from a semantic map. These elements serve as a multi-source knowledge input—encompassing commonsense, tool, and spatial knowledge—to aid the model's understanding of the scene. We have collected and organized all of this information for the CookBench environment.

The agent receives dynamic \textbf{Perceptual Feedback} through its ``eyes''. 
This involves using a Vision-Language Model(VLM) to continuously analyze RGB images from the environment, providing feedback on the execution status of its actions (e.g., success or failure) and detecting unexpected situations (e.g., a dropped item). This ensures that the agent's plan can be adjusted and corrected according to the actual situation.

The evaluation metrics for this stage include the \textbf{Task Completion Score} and \textbf{Completion Time}. After the agent submits the final dish, the simulation environment performs an automated evaluation. It objectively assesses various factors, including the correctness of ingredients, the application of proper cooking techniques, and the final state of components (e.g., temperature), to generate a completion score on a five-point scale. The system also provides detailed textual feedback for specific errors, such as \textit{lemon: Bad cutting technique}, indicating that the lemon was not cut as required. 

\begin{table*}[h!]
\centering
\begin{tabular}{@{} ll cccc cccc c @{}} 
\toprule
\multicolumn{2}{c}{} & \multicolumn{2}{c}{Single-Dish Instruction} & \multicolumn{6}{c}{Multi-Dish Instruction} & \multirow{3}{*}{Average} \\
\cmidrule(lr){3-4} \cmidrule(lr){5-10}

\multicolumn{2}{c}{\textbf{Model(\%)}} & \multirow{2}{*}{Chinese} & \multirow{2}{*}{English} & \multicolumn{2}{c}{4-Dish} & \multicolumn{2}{c}{5-Dish} & \multicolumn{2}{c}{6-Dish} & \\
\cmidrule(lr){5-6} \cmidrule(lr){7-8} \cmidrule(lr){9-10}

\multicolumn{2}{c}{} & & & Chinese & English & Chinese & English & Chinese & English & \\
\midrule
\multirow{7}{*}{\shortstack{Open-\\Source}} & Qwen3-235B      &79.55 &93.17 &34.73 &52.40 &38.68 &52.68 &29.97 &44.95 &59.11 \\
& Qwen3-32B       &76.15 &90.88 &29.96 &42.69 &27.12 &38.27 &24.47 &31.65 &51.91 \\
& Qwen3-8B       &72.88 &86.22 &23.08 &25.30 &16.51 &20.57 &14.61 &17.11 &42.47 \\
& Qwen3-1.7B        &51.58 &73.90 &1.69 &2.16 &1.07 &1.54 &0.41 &0.82 &24.76 \\
& Qwen3-0.6B      &52.53 &70.81 &0.27 &1.76 &0.40 &1.07 &0.00 &0.14 &23.90 \\
& DeepSeek-V3    &82.44 &94.15 &18.29 &26.79 &11.62 &17.10 &7.77 &6.95 &42.87 \\
& DeepSeek-R1    &90.77 &94.58 &46.98 &58.81 &44.13 &56.78 &38.18 &49.83 &65.78 \\
\midrule
\multirow{2}{*}{\shortstack{Closed-\\Source}} & Gpt-4.1        &87.57 &96.69 &42.71 &57.22 &31.66 &48.36 &25.49 &37.29 &60.22 \\
& Gemini-2.5-pro &92.04 &96.44 &61.54 &67.61 &59.66 &66.13 &47.65 &55.56 &72.91 \\
\bottomrule
\end{tabular}
\caption{Intent recognition accuracy (\%) of various LLMs on the CookBench-IR dataset. To comprehensively test model generalization, the evaluation set includes a diverse mix of single-turn and multi-turn dialogues, batch and sequential orders, as well as various dish repetition patterns (all-same, all-different, and partially-same). To realistically simulate real-world conditions, the 4, 5, and 6-dish instructions are shuffled together during evaluation.}
\label{tab:final_with_average}
\end{table*}

\section{Experiments and Analysis}
\label{sec:experiments}

The tasks in \textbf{CookBench} inherently follow a two-stage process: first, an Intention Recognition stage, where the agent should parse natural language instructions into structured cooking goals; and second, an Embodied Execution stage, where the agent should perform long-horizon physical interactions in the simulator based on these goals. To systematically evaluate the capability boundaries of current SOTA models against the challenges of each stage, we have designed a series of decoupled experiments.

\subsection{Stage 1: Performance on Intention Recognition}

Our first stage of experiments investigates the capabilities of current LLMs on the complex task of intent recognition. We assessed a range of language models, including both open and closed-source variants of various sizes, testing them on our bilingual (Chinese/English) instruction sets via API calls. The core task is to predict a final list of target dish(es) from an instruction, which should also include the correct sequence for multi-dish order(s). We employ a strict accuracy metric: a prediction is considered correct only if all dish name(s) and their order exactly match the ground truth.


\paragraph{Experimental Setup.}
We assess a range of language models, including both open and closed-source variants of various sizes. We tested these models, including top-tier models like \texttt{Gpt-4.1}~\cite{achiam2023gpt}, \texttt{Gemini-2.5-pro}~\cite{comanici2025gemini}, \texttt{DeepSeek-R1}~\cite{guo2025deepseek}, and the \texttt{Qwen3} family~\cite{yang2025qwen3}, on our bilingual (Chinese/English) instruction sets via their respective APIs.

The results are summarized in Table~\ref{tab:final_with_average}. The leading closed-source and open-source models all achieve excellent performance, with accuracies exceeding 90\% on the English set, suggesting that providing a reliable goal for downstream embodied tasks is feasible. However, performance is strongly correlated with model size. The \texttt{Qwen3} family, for instance, shows an accuracy gap of over 20 percentage points between its largest and smallest variants.

As tasks shift to the more complex multi-dish instructions, the performance of all models drops precipitously. Even the top-performing model, gemini-2.5-pro, sees its accuracy fall to around 50\% on 6-dish tasks, while smaller-parameter models effectively fail. We also observe that accuracy on the Chinese dataset is consistently lower than on the English version, which corresponds with our design choice to intentionally increase its difficulty using aliases and greater semantic complexity. These findings not only validate the challenging nature of our CookBench-IR dataset but also highlight that even leading large models face significant planning and reasoning bottlenecks at the very initial stage of a complex embodied task.
\subsection{Stage 2: A Human-in-the-Loop Feasibility Study}



\begin{table*}[h!]
\centering
\begin{tabular*}{\textwidth}{@{\extracolsep{\fill}}lcccccc}
\toprule
\textbf{Method} & \makecell{French \& Mediterranean} & \makecell{American} & \makecell{Central/Eastern \& \\Other European} & Italian & \makecell{Asian \& Others} & Chinese \\
\midrule
HITL             & 0.49 & 0.43 & 0.33 & 0.32 & 0.70 & 0.75 \\
Human           & 5 & 5 & 5 & 5 & 5 & 5 \\
\bottomrule
\end{tabular*}
\caption{A performance comparison between our Human-in-the-Loop (HITL) agent and human experts on the 131 single-dish tasks, with scores averaged by cuisine style. The Human row represents the expert performance level; the perfect score of 5 reflects that two volunteers achieved the maximum score on all tasks during interaction.}
\label{tab:human_performance}
\end{table*}

Our second set of experiments evaluates the Embodied Interaction stage. To isolate this stage's challenges from intention recognition errors, agents were provided with the ground truth cooking intent. The task set consisted of 131 single-dish and 4,446 multi-dish combinations.

\begin{figure}[t]
  \centering
  \includegraphics[width=\columnwidth]{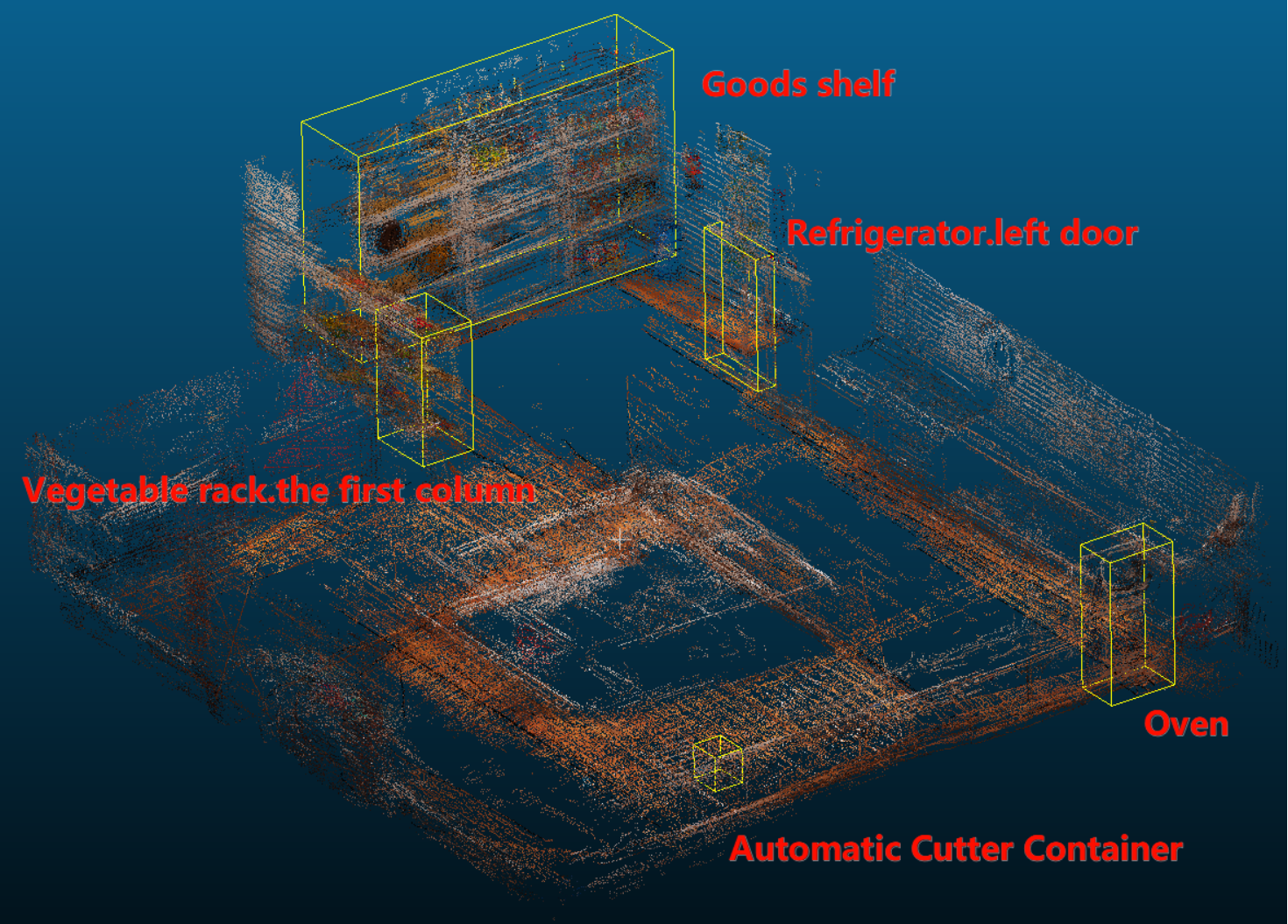}
  \caption{3D semantic map of the Cookbench environment. This map contains over 70 manually annotated semantic categories. For clarity, the figure highlights five of them, illustrating a foundational map that helps models understand the semantic information and spatial relationships of the scene.}
  \label{fig:semantic_map}
\end{figure}

\paragraph{Experimental Setup.}
Our experimental agent's architecture combines several components: we select \texttt{gpt-4.1} as the high-level planner responsible for decomposing recipes into sub-tasks, \texttt{gpt-4o} as a perception model that also participates in some planning. To supplement these models, we also provide the agent with a manually curated database of scene object and tool information (including weights, initial relative positions, and supported operations), as well as our spatial knowledge graph derived from a 3D semantic map shown in Figure~\ref{fig:semantic_map}.

\begin{figure}[t]
  \centering
  \includegraphics[width=\columnwidth]{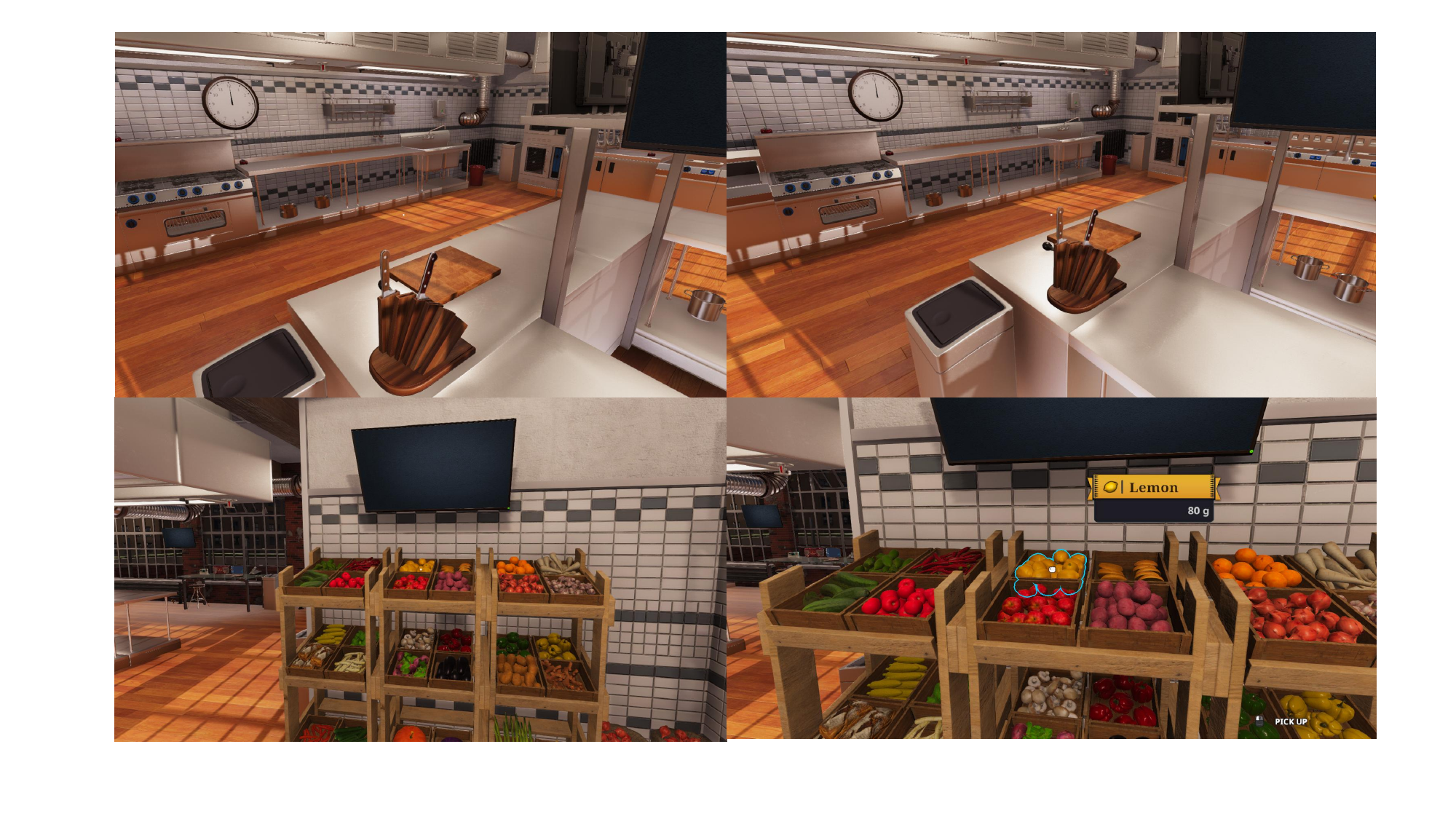}
  \caption{An illustration of two common planning failures in CookBench. Top: An agent gets stuck in a corner; recovery requires a manually-corrected command, ``move\_back(0.5)'', to retreat. Bottom: An agent fails to pick up an item from too far away; the corrective action is a manually-provided ``move\_forward(0.3)'' command to first close the distance.}
  \label{fig:failure_case}
\end{figure}

\paragraph{Feasibility Study with Human-in-the-Loop.}
CookBench defines a comprehensive challenge that intersects vision, navigation, and manipulation, for which no current model can currently provide a fully autonomous solution. Therefore, this study aims to take a critical first step: we use a Human-in-the-Loop (HITL) approach to specifically evaluate the high-level task decomposition and logical reasoning capabilities of leading planning models, while explicitly positing full automation as a direction for future research.The results are shown in Table~\ref{tab:human_performance}.

Taking the Easy category task ``Lemon Tart'' as an example, the method involves the model performing the initial high-level planning, with a human assisting in the implementation of specific low-level details to complete the task. To minimize intervention at the planning level, human assistance was restricted to eight atomic commands: four for navigation and four for camera rotation.

First, the planning model successfully decomposed the ``Lemon Tart'' task into a high-level sequence of sub-tasks, such as \textit{place order, take Tart [260g], bake Tart for 40s, add Coconut Milk [180ml] to Food Processor, add Honey [20ml] to Food Processor, add Brown Sugar [5g] to Food Processor, blend mixture in Food Processor, cut Banana [100g] into Pieces [20g], cut Lemon [80g] into Pieces [10g], cut Orange [60g] into Quarters [15g], cut Strawberry [60g] into Halves [5g], transfer Mixture [200ml] into Tart, add Lemon Pieces to Tart, add Banana Pieces to Tart, add Orange Quarters to Tart, add Strawberry Halves to Tart, serve Tart cool, submit dish}.

However, at the concrete navigation level, both gpt-4.1 and gpt-4o failed, frequently getting stuck in corners or attempting to interact with objects from a distance, as shown in Figure~\ref{fig:failure_case}. Therefore, navigation and the precise alignment for object pickup were accomplished through human correction of atomic action commands.
This process enabled the agent to navigate and retrieve the Tart and the lemon. Although the process was manually terminated at later complex stages due to planning loops, the submitted semi-finished dish still received a 2/5 score from the automated evaluation. This demonstrates the usability of CookBench as a testbed.

\paragraph{Key Challenges for Current Models.}
Our study reveals several core challenges that should be overcome to implement a powerful model in CookBench, defining clear avenues for future research:

Multi-Source Reasoning for Planning: The agent should synthesize information from disparate sources. For example, it must combine recipe knowledge (a Tart is needed) with state knowledge (the Tart is missing from the scene) and spatial knowledge (the shop is located at the 'first shelf column') to deduce the sub-goal of purchasing the item.

Vision-based Semantic Navigation: Planning models struggle to generate effective navigation action sequences based solely on visual input and implicit task rules.

Visual Reasoning \& Latency: The latency of VLMs limits their application in tasks requiring real-time responsiveness, and they struggle to visually verify action outcomes or detect physical anomalies (like being stuck).

Fine-grained Spatial Interaction: For functional spatial operations like slicing a lemon with a knife, current VLMs cannot yet generate the required precise spatial parameters.


\paragraph{Spatial Knowledge Graph from 3D Semantic Maps.} 
\label{sec:sem_map}
To aid the agent in spatial reasoning, we construct a spatial knowledge graph.  The construction process begins with RGB images captured from the environment, which are first used for 3D point cloud reconstruction via the VGGT model~\cite{wang2025vggt}. Subsequently, the reconstructed point cloud undergoes a manual process of object classification and semantic annotation. This process yields precise 3D bounding boxes for each object, including their spatial extent, position, and semantic label. The annotation is fine-grained enough to specify the locations of sub-parts, such as the first column of the vegetable rack, as shown in Figure~\ref{fig:semantic_map}. Based on this detailed annotation, we construct the spatial knowledge graph. In this graph, nodes represent objects and their designated sub-parts, while the edges represent their relative spatial relationships at various levels, such as object-to-object or sub-part-to-object. Furthermore, the graph can provide quantifiable distance and size information derived from the scale of the reconstructed point cloud.

\bibliography{aaai2026}

\begin{thebibliography}{42}
\providecommand{\natexlab}[1]{#1}

\bibitem[{Achiam et~al.(2023)Achiam, Adler, Agarwal, Ahmad, Akkaya, Aleman, Almeida, Altenschmidt, Altman, Anadkat et~al.}]{achiam2023gpt}
Achiam, J.; Adler, S.; Agarwal, S.; Ahmad, L.; Akkaya, I.; Aleman, F.~L.; Almeida, D.; Altenschmidt, J.; Altman, S.; Anadkat, S.; et~al. 2023.
\newblock Gpt-4 technical report.
\newblock \emph{arXiv preprint arXiv:2303.08774}.

\bibitem[{Bollini et~al.(2013)Bollini, Tellex, Thompson, Roy, and Rus}]{bollini2013interpreting}
Bollini, M.; Tellex, S.; Thompson, T.; Roy, N.; and Rus, D. 2013.
\newblock Interpreting and executing recipes with a cooking robot.
\newblock In \emph{Experimental Robotics: The 13th International Symposium on Experimental Robotics}, 481--495. Springer.

\bibitem[{Chen et~al.(2024)Chen, Xu, Kirmani, Ichter, Sadigh, Guibas, and Xia}]{chen2024spatialvlm}
Chen, B.; Xu, Z.; Kirmani, S.; Ichter, B.; Sadigh, D.; Guibas, L.; and Xia, F. 2024.
\newblock Spatialvlm: Endowing vision-language models with spatial reasoning capabilities.
\newblock In \emph{Proceedings of the IEEE/CVF Conference on Computer Vision and Pattern Recognition}, 14455--14465.

\bibitem[{Comanici et~al.(2025)Comanici, Bieber, Schaekermann, Pasupat, Sachdeva, Dhillon, Blistein, Ram, Zhang, Rosen et~al.}]{comanici2025gemini}
Comanici, G.; Bieber, E.; Schaekermann, M.; Pasupat, I.; Sachdeva, N.; Dhillon, I.; Blistein, M.; Ram, O.; Zhang, D.; Rosen, E.; et~al. 2025.
\newblock Gemini 2.5: Pushing the frontier with advanced reasoning, multimodality, long context, and next generation agentic capabilities.
\newblock \emph{arXiv preprint arXiv:2507.06261}.

\bibitem[{Deitke et~al.(2022)Deitke, Vander~Bilt, Herrasti, Weihs, Salvador, Ehsani, Han, Kolve, Farhadi, Kembhavi et~al.}]{deitke2022procthor}
Deitke, M.; Vander~Bilt, E.; Herrasti, A.; Weihs, L.; Salvador, J.; Ehsani, K.; Han, W.; Kolve, E.; Farhadi, A.; Kembhavi, A.; et~al. 2022.
\newblock ProcTHOR: large-scale embodied AI using procedural generation.
\newblock In \emph{Proceedings of the 36th International Conference on Neural Information Processing Systems}, 5982--5994.

\bibitem[{Duan et~al.(2022)Duan, Yu, Tan, Zhu, and Tan}]{duan2022survey}
Duan, J.; Yu, S.; Tan, H.~L.; Zhu, H.; and Tan, C. 2022.
\newblock A survey of embodied ai: From simulators to research tasks.
\newblock \emph{IEEE Transactions on Emerging Topics in Computational Intelligence}, 6(2): 230--244.

\bibitem[{Feldotto et~al.(2022)Feldotto, Eppler, Jimenez-Romero, Bignamini, Gutierrez, Albanese, Retamino, Vorobev, Zolfaghari, Upton et~al.}]{feldotto2022deploying}
Feldotto, B.; Eppler, J.~M.; Jimenez-Romero, C.; Bignamini, C.; Gutierrez, C.~E.; Albanese, U.; Retamino, E.; Vorobev, V.; Zolfaghari, V.; Upton, A.; et~al. 2022.
\newblock Deploying and optimizing embodied simulations of large-scale spiking neural networks on HPC infrastructure.
\newblock \emph{Frontiers in neuroinformatics}, 16: 884180.

\bibitem[{Gu et~al.(2023)Gu, Xiang, Li, Ling, Liu, Mu, Tang, Tao, Wei, Yao et~al.}]{gu2023maniskill2}
Gu, J.; Xiang, F.; Li, X.; Ling, Z.; Liu, X.; Mu, T.; Tang, Y.; Tao, S.; Wei, X.; Yao, Y.; et~al. 2023.
\newblock Maniskill2: A unified benchmark for generalizable manipulation skills.
\newblock \emph{arXiv preprint arXiv:2302.04659}.

\bibitem[{Guo et~al.(2025)Guo, Yang, Zhang, Song, Zhang, Xu, Zhu, Ma, Wang, Bi et~al.}]{guo2025deepseek}
Guo, D.; Yang, D.; Zhang, H.; Song, J.; Zhang, R.; Xu, R.; Zhu, Q.; Ma, S.; Wang, P.; Bi, X.; et~al. 2025.
\newblock Deepseek-r1: Incentivizing reasoning capability in llms via reinforcement learning.
\newblock \emph{arXiv preprint arXiv:2501.12948}.

\bibitem[{Haas(2014)}]{haas2014history}
Haas, J.~K. 2014.
\newblock A history of the unity game engine.

\bibitem[{Huang et~al.(2022)Huang, Xia, Xiao, Chan, Liang, Florence, Zeng, Tompson, Mordatch, Chebotar et~al.}]{huang2022inner}
Huang, W.; Xia, F.; Xiao, T.; Chan, H.; Liang, J.; Florence, P.; Zeng, A.; Tompson, J.; Mordatch, I.; Chebotar, Y.; et~al. 2022.
\newblock Inner monologue: Embodied reasoning through planning with language models.
\newblock \emph{arXiv preprint arXiv:2207.05608}.

\bibitem[{Ivanova et~al.(2025)Ivanova, Bakaeva, Volovikova, Kovalev, and Panov}]{ivanova2025ambik}
Ivanova, A.; Bakaeva, E.; Volovikova, Z.; Kovalev, A.~K.; and Panov, A.~I. 2025.
\newblock AmbiK: Dataset of Ambiguous Tasks in Kitchen Environment.
\newblock \emph{arXiv preprint arXiv:2506.04089}.

\bibitem[{Jin et~al.(2023)Jin, Hu, Huang, Zhang, Wu, Fei-Fei, and Mart{\'\i}n-Mart{\'\i}n}]{jin2023mini}
Jin, E.; Hu, J.; Huang, Z.; Zhang, R.; Wu, J.; Fei-Fei, L.; and Mart{\'\i}n-Mart{\'\i}n, R. 2023.
\newblock Mini-behavior: A procedurally generated benchmark for long-horizon decision-making in embodied ai.
\newblock \emph{arXiv preprint arXiv:2310.01824}.

\bibitem[{Kanazawa et~al.(2024)Kanazawa, Kawaharazuka, Obinata, Okada, and Inaba}]{kanazawa2024real}
Kanazawa, N.; Kawaharazuka, K.; Obinata, Y.; Okada, K.; and Inaba, M. 2024.
\newblock Real-world cooking robot system from recipes based on food state recognition using foundation models and PDDL.
\newblock \emph{Advanced Robotics}, 38(18): 1318--1334.

\bibitem[{Kaur, Singh, and Banerjee(2023)}]{kaur2023review}
Kaur, D.~P.; Singh, N.~P.; and Banerjee, B. 2023.
\newblock A review of platforms for simulating embodied agents in 3D virtual environments.
\newblock \emph{Artificial Intelligence Review}, 56(4): 3711--3753.

\bibitem[{Kim et~al.(2024)Kim, Min, Kim, Kim, Jeung, and Choi}]{kim2024realfred}
Kim, T.; Min, C.; Kim, B.; Kim, J.; Jeung, W.; and Choi, J. 2024.
\newblock ReALFRED: An Embodied Instruction Following Benchmark in Photo-Realistic Environments.
\newblock In \emph{European Conference on Computer Vision}, 346--364. Springer.

\bibitem[{Li et~al.(2021)Li, Xia, Mart{\'\i}n-Mart{\'\i}n, Lingelbach, Srivastava, Shen, Vainio, Gokmen, Dharan, Jain et~al.}]{li2021igibson}
Li, C.; Xia, F.; Mart{\'\i}n-Mart{\'\i}n, R.; Lingelbach, M.; Srivastava, S.; Shen, B.; Vainio, K.; Gokmen, C.; Dharan, G.; Jain, T.; et~al. 2021.
\newblock igibson 2.0: Object-centric simulation for robot learning of everyday household tasks.
\newblock \emph{arXiv preprint arXiv:2108.03272}.

\bibitem[{Li et~al.(2025)Li, Wang, Xu, Ye, and Chen}]{li2025developments}
Li, G.; Wang, R.; Xu, P.; Ye, Q.; and Chen, J. 2025.
\newblock The Developments and Challenges towards Dexterous and Embodied Robotic Manipulation: A Survey.
\newblock \emph{arXiv preprint arXiv:2507.11840}.

\bibitem[{Liang et~al.(2022)Liang, Huang, Xia, Xu, Hausman, Ichter, Florence, and Zeng}]{liang2022code}
Liang, J.; Huang, W.; Xia, F.; Xu, P.; Hausman, K.; Ichter, B.; Florence, P.; and Zeng, A. 2022.
\newblock Code as policies: Language model programs for embodied control.
\newblock \emph{arXiv preprint arXiv:2209.07753}.

\bibitem[{Lin et~al.(2025)Lin, Huang, Li, Jiang, Wu, Zhong, Qian, Wang, and Qi}]{lin2025embrace}
Lin, M.; Huang, W.; Li, Y.; Jiang, C.; Wu, K.; Zhong, F.; Qian, S.; Wang, X.; and Qi, X. 2025.
\newblock EmbRACE-3K: Embodied Reasoning and Action in Complex Environments.
\newblock \emph{arXiv preprint arXiv:2507.10548}.

\bibitem[{Liu, Guo, and Cangelosi(2025)}]{liu2025embodied}
Liu, H.; Guo, D.; and Cangelosi, A. 2025.
\newblock Embodied intelligence: A synergy of morphology, action, perception and learning.
\newblock \emph{ACM Computing Surveys}, 57(7): 1--36.

\bibitem[{Liu et~al.(2024)Liu, Chen, Bai, Liang, Li, Gao, and Lin}]{liu2024aligning}
Liu, Y.; Chen, W.; Bai, Y.; Liang, X.; Li, G.; Gao, W.; and Lin, L. 2024.
\newblock Aligning cyber space with physical world: A comprehensive survey on embodied ai.
\newblock \emph{arXiv preprint arXiv:2407.06886}.

\bibitem[{Mavrogiannis, Mavrogiannis, and Aloimonos(2024)}]{mavrogiannis2024cook2ltl}
Mavrogiannis, A.; Mavrogiannis, C.; and Aloimonos, Y. 2024.
\newblock Cook2ltl: Translating cooking recipes to ltl formulae using large language models.
\newblock In \emph{2024 IEEE International Conference on Robotics and Automation (ICRA)}, 17679--17686. IEEE.

\bibitem[{Padmakumar et~al.(2022)Padmakumar, Thomason, Shrivastava, Lange, Narayan-Chen, Gella, Piramuthu, Tur, and Hakkani-Tur}]{padmakumar2022teach}
Padmakumar, A.; Thomason, J.; Shrivastava, A.; Lange, P.; Narayan-Chen, A.; Gella, S.; Piramuthu, R.; Tur, G.; and Hakkani-Tur, D. 2022.
\newblock Teach: Task-driven embodied agents that chat.
\newblock In \emph{Proceedings of the AAAI Conference on Artificial Intelligence}, volume~36, 2017--2025.

\bibitem[{Puig et~al.(2018)Puig, Ra, Boben, Li, Wang, Fidler, and Torralba}]{puig2018virtualhome}
Puig, X.; Ra, K.; Boben, M.; Li, J.; Wang, T.; Fidler, S.; and Torralba, A. 2018.
\newblock Virtualhome: Simulating household activities via programs.
\newblock In \emph{Proceedings of the IEEE conference on computer vision and pattern recognition}, 8494--8502.

\bibitem[{Shridhar et~al.(2020{\natexlab{a}})Shridhar, Thomason, Gordon, Bisk, Han, Mottaghi, Zettlemoyer, and Fox}]{shridhar2020alfred}
Shridhar, M.; Thomason, J.; Gordon, D.; Bisk, Y.; Han, W.; Mottaghi, R.; Zettlemoyer, L.; and Fox, D. 2020{\natexlab{a}}.
\newblock Alfred: A benchmark for interpreting grounded instructions for everyday tasks.
\newblock In \emph{Proceedings of the IEEE/CVF conference on computer vision and pattern recognition}, 10740--10749.

\bibitem[{Shridhar et~al.(2020{\natexlab{b}})Shridhar, Yuan, C{\^o}t{\'e}, Bisk, Trischler, and Hausknecht}]{shridhar2020alfworld}
Shridhar, M.; Yuan, X.; C{\^o}t{\'e}, M.-A.; Bisk, Y.; Trischler, A.; and Hausknecht, M. 2020{\natexlab{b}}.
\newblock Alfworld: Aligning text and embodied environments for interactive learning.
\newblock \emph{arXiv preprint arXiv:2010.03768}.

\bibitem[{Song et~al.(2023)Song, Wu, Washington, Sadler, Chao, and Su}]{song2023llm}
Song, C.~H.; Wu, J.; Washington, C.; Sadler, B.~M.; Chao, W.-L.; and Su, Y. 2023.
\newblock Llm-planner: Few-shot grounded planning for embodied agents with large language models.
\newblock In \emph{Proceedings of the IEEE/CVF international conference on computer vision}, 2998--3009.

\bibitem[{Srivastava et~al.(2022)Srivastava, Li, Lingelbach, Mart{\'\i}n-Mart{\'\i}n, Xia, Vainio, Lian, Gokmen, Buch, Liu et~al.}]{srivastava2022behavior}
Srivastava, S.; Li, C.; Lingelbach, M.; Mart{\'\i}n-Mart{\'\i}n, R.; Xia, F.; Vainio, K.~E.; Lian, Z.; Gokmen, C.; Buch, S.; Liu, K.; et~al. 2022.
\newblock Behavior: Benchmark for everyday household activities in virtual, interactive, and ecological environments.
\newblock In \emph{Conference on robot learning}, 477--490. PMLR.

\bibitem[{Todorov, Erez, and Tassa(2012)}]{todorov2012mujoco}
Todorov, E.; Erez, T.; and Tassa, Y. 2012.
\newblock Mujoco: A physics engine for model-based control.
\newblock In \emph{2012 IEEE/RSJ international conference on intelligent robots and systems}, 5026--5033. IEEE.

\bibitem[{Wang et~al.(2025)Wang, Chen, Karaev, Vedaldi, Rupprecht, and Novotny}]{wang2025vggt}
Wang, J.; Chen, M.; Karaev, N.; Vedaldi, A.; Rupprecht, C.; and Novotny, D. 2025.
\newblock Vggt: Visual geometry grounded transformer.
\newblock In \emph{Proceedings of the Computer Vision and Pattern Recognition Conference}, 5294--5306.

\bibitem[{Wu et~al.(2023)Wu, Wang, Xu, Lu, and Yan}]{wu2023embodied}
Wu, Z.; Wang, Z.; Xu, X.; Lu, J.; and Yan, H. 2023.
\newblock Embodied task planning with large language models.
\newblock \emph{arXiv preprint arXiv:2307.01848}.

\bibitem[{Xiang et~al.(2020)Xiang, Qin, Mo, Xia, Zhu, Liu, Liu, Jiang, Yuan, Wang et~al.}]{xiang2020sapien}
Xiang, F.; Qin, Y.; Mo, K.; Xia, Y.; Zhu, H.; Liu, F.; Liu, M.; Jiang, H.; Yuan, Y.; Wang, H.; et~al. 2020.
\newblock Sapien: A simulated part-based interactive environment.
\newblock In \emph{Proceedings of the IEEE/CVF conference on computer vision and pattern recognition}, 11097--11107.

\bibitem[{Xu et~al.(2025)Xu, Gao, Yu, An, Chen, Wang, Guo, Liang, and Xu}]{xu20253d}
Xu, R.; Gao, H.; Yu, M.; An, D.; Chen, S.; Wang, C.; Guo, L.; Liang, X.; and Xu, S. 2025.
\newblock 3D-MoRe: Unified Modal-Contextual Reasoning for Embodied Question Answering.
\newblock \emph{arXiv preprint arXiv:2507.12026}.

\bibitem[{Yang et~al.(2025{\natexlab{a}})Yang, Li, Yang, Zhang, Hui, Zheng, Yu, Gao, Huang, Lv et~al.}]{yang2025qwen3}
Yang, A.; Li, A.; Yang, B.; Zhang, B.; Hui, B.; Zheng, B.; Yu, B.; Gao, C.; Huang, C.; Lv, C.; et~al. 2025{\natexlab{a}}.
\newblock Qwen3 technical report.
\newblock \emph{arXiv preprint arXiv:2505.09388}.

\bibitem[{Yang et~al.(2025{\natexlab{b}})Yang, Tan, Wu, Zheng, Peng, Liang, Gu, Cai, Ye, Jang et~al.}]{yang2025magma}
Yang, J.; Tan, R.; Wu, Q.; Zheng, R.; Peng, B.; Liang, Y.; Gu, Y.; Cai, M.; Ye, S.; Jang, J.; et~al. 2025{\natexlab{b}}.
\newblock Magma: A foundation model for multimodal ai agents.
\newblock In \emph{Proceedings of the Computer Vision and Pattern Recognition Conference}, 14203--14214.

\bibitem[{Yang et~al.(2025{\natexlab{c}})Yang, Chen, Zhang, Zhao, Qian, Wang, Wang, Koripella, Movahedi, Li et~al.}]{yang2025embodiedbench}
Yang, R.; Chen, H.; Zhang, J.; Zhao, M.; Qian, C.; Wang, K.; Wang, Q.; Koripella, T.~V.; Movahedi, M.; Li, M.; et~al. 2025{\natexlab{c}}.
\newblock Embodiedbench: Comprehensive benchmarking multi-modal large language models for vision-driven embodied agents.
\newblock \emph{arXiv preprint arXiv:2502.09560}.

\bibitem[{Yang et~al.(2021)Yang, Ji, Wu, and Lai}]{yang2021open}
Yang, X.; Ji, Z.; Wu, J.; and Lai, Y.-K. 2021.
\newblock An open-source multi-goal reinforcement learning environment for robotic manipulation with pybullet.
\newblock In \emph{Annual Conference Towards Autonomous Robotic Systems}, 14--24. Springer.

\bibitem[{Zhang et~al.(2025)Zhang, Yun, Cen, Cai, Zhu, Yuan, Zhao, Feng, Wang, Chen et~al.}]{zhang2025generative}
Zhang, K.; Yun, P.; Cen, J.; Cai, J.; Zhu, D.; Yuan, H.; Zhao, C.; Feng, T.; Wang, M.~Y.; Chen, Q.; et~al. 2025.
\newblock Generative artificial intelligence in robotic manipulation: A survey.
\newblock \emph{arXiv preprint arXiv:2503.03464}.

\bibitem[{Zhang et~al.(2024)Zhang, Wang, Gu, Hamidizadeh, Zhang, Liu, Wang, Bravo, Dong, Zhou et~al.}]{zhang2024plan}
Zhang, L.; Wang, Y.; Gu, H.; Hamidizadeh, A.; Zhang, Z.; Liu, Y.; Wang, Y.; Bravo, D. G.~A.; Dong, J.; Zhou, S.; et~al. 2024.
\newblock Et-plan-bench: Embodied task-level planning benchmark towards spatial-temporal cognition with foundation models.
\newblock \emph{arXiv preprint arXiv:2410.14682}.

\bibitem[{Zhou et~al.(2025)Zhou, An, Chi, Han, Rong, Zhang, Wang, Wang, Huang, Sheng et~al.}]{zhou2025roborefer}
Zhou, E.; An, J.; Chi, C.; Han, Y.; Rong, S.; Zhang, C.; Wang, P.; Wang, Z.; Huang, T.; Sheng, L.; et~al. 2025.
\newblock RoboRefer: Towards Spatial Referring with Reasoning in Vision-Language Models for Robotics.
\newblock \emph{arXiv preprint arXiv:2506.04308}.

\bibitem[{Zhu et~al.(2025)Zhu, Wang, Li, Zhang, Ma, Chen, Jia, Liang, Yu, Deng et~al.}]{zhu2025move}
Zhu, Z.; Wang, X.; Li, Y.; Zhang, Z.; Ma, X.; Chen, Y.; Jia, B.; Liang, W.; Yu, Q.; Deng, Z.; et~al. 2025.
\newblock Move to Understand a 3D Scene: Bridging Visual Grounding and Exploration for Efficient and Versatile Embodied Navigation.
\newblock \emph{arXiv preprint arXiv:2507.04047}.

\end{thebibliography}

\end{document}